\documentclass[10pt,twocolumn,letterpaper]{article}

\usepackage{cvpr}              % To produce the CAMERA-READY version
\usepackage[dvipsnames]{xcolor}

\newcommand{\fnote}[1]{}

\newcommand{\kcut}[1]{}
\newcommand{\scut}[1]{}
\newcommand{\ocut}[1]{}

\definecolor{cvprblue}{rgb}{0.21,0.49,0.74}
\usepackage[pagebackref,breaklinks,colorlinks,citecolor=cvprblue]{hyperref}
\usepackage{multirow}

\def\paperID{203} % *** Enter the Paper ID here
\def\confName{3DV\xspace}
\def\confYear{2024\xspace}

\title{Depth Reconstruction with Neural Signed Distance Fields in Structured Light Systems}

\author{Rukun Qiao\\
Peking University, China\\
{\tt\small rukunqiao@pku.edu.cn}
\and
Hiroshi Kawasaki\\
Kyushu University, Japan\\
{\tt\small kawasaki@ait.kyushu-u.ac.jp}
\and
Hongbin Zha\\
Peking University, China\\
{\tt\small zha@cis.pku.edu.cn}
}
\begin{document}
\maketitle

\begin{abstract}
% We present a novel depth estimation method for multi-frame structured light systems via the neural implicit representations of the 3D space. Specifically, we represent the 3D space with a neural signed distance field (SDF), which is trained using a differentiable rendering process in a self-supervised manner. In contrast to the passive vision case where radiance fields and geometry fields has to be estimated jointly, we leverage the projected patterns in structured light systems as the prior known radiance field. Therefore, we can optimize the geometry field solely, ensuring the convergence and performance of our neural network given a fixed device position. To further ensure a clear and smooth geometric output from the neural function, we apply an additional color loss based on the object surface during the training process. Experiments on the real-world data show that our method not only outperforms the state-of-the-arts in terms of geometric performance for few-shot cases, but also achieve comparable results when more patterns are available.
We introduce a novel depth estimation technique for multi-frame structured light setups using neural implicit representations of 3D space. Our approach employs a neural signed distance field (SDF), trained through self-supervised differentiable rendering. Unlike passive vision, where joint estimation of radiance and geometry fields is necessary, we capitalize on known radiance fields from projected patterns in structured light systems. This enables isolated optimization of the geometry field, ensuring convergence and network efficacy with fixed device positioning. To enhance geometric fidelity, we incorporate an additional color loss based on object surfaces during training. Real-world experiments demonstrate our method's superiority in geometric performance for few-shot scenarios, while achieving comparable results with increased pattern availability.
\end{abstract}

% \textbf{Need to be improved according to Introduction. This time we want to emphasize the learning framework without dataset, and constraints from pattern makes geometry better. In the mean time, try to limit the topic in few-shot case.} We present a novel depth estimation method for structured light systems via the neural implicit representations of the object surface. Inspired by the Neural Radiance Fields (NeRF) and related works in the view synthesis problem, we represent the 3D space with a neural signed distance field (SDF). This neural field is trained using a differentiable rendering process in a self-supervised manner. In contrast to the passive vision case, we leverage the projected patterns to improve the geometric performance and ensure convergence of our network given a fixed device position. We achieve this by proposing a physical rendering model that includes the projected patterns to render the images in structured light systems. To ensure a clear and smooth geometric output from the neural function, we apply a warping loss, constrained by the projected patterns, during the training process. Experiments on the real-world data show that our method outperforms the state-of-the-arts in terms of geometric performance.

\section{Introduction}

\begin{figure}
    \centering
    \includegraphics[width=\linewidth]{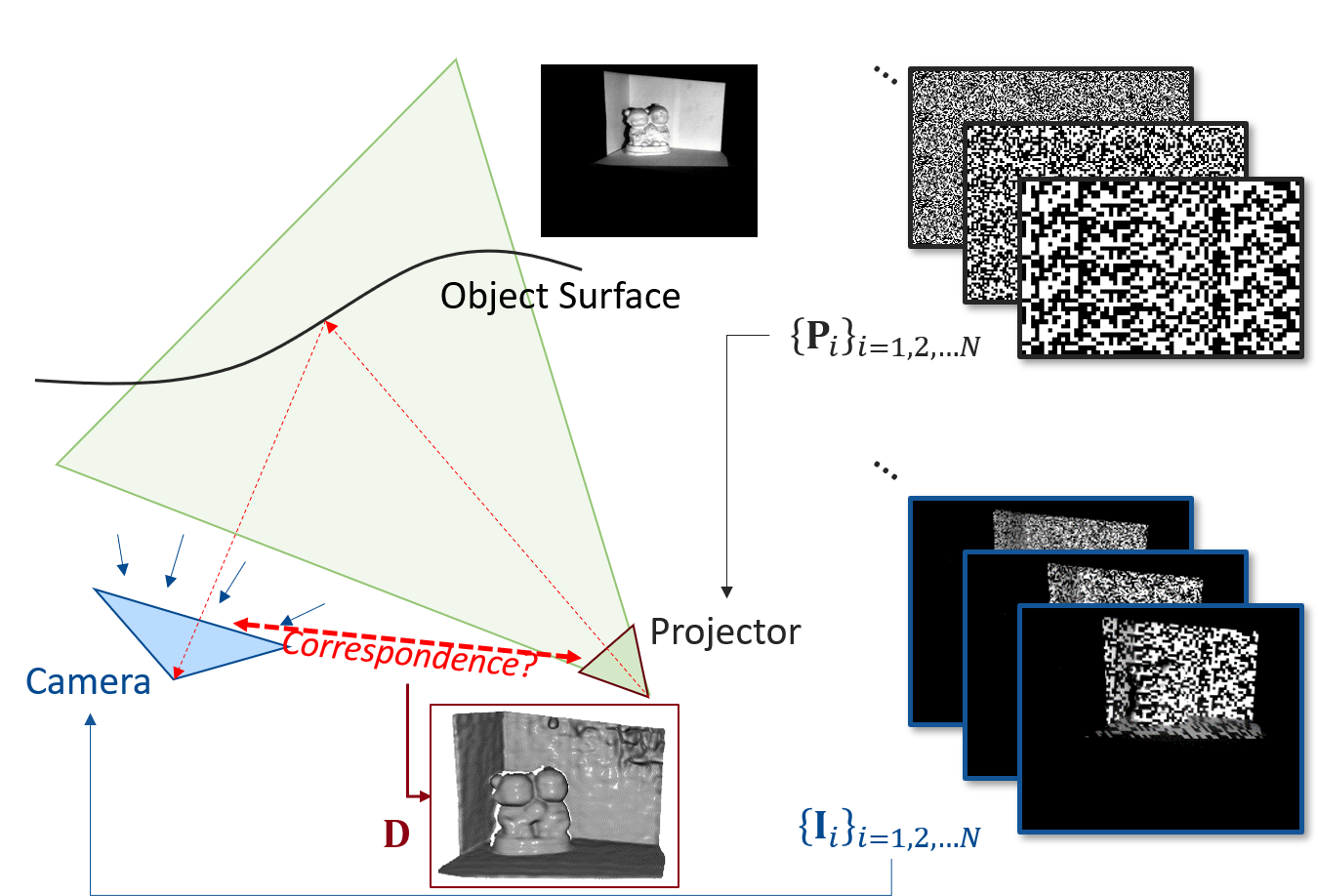}
    \caption{A schematic diagram of a monocular structured light system. Once a correspondence between the camera and the projector is established, the depth information can be computed.
    % \knote{can you make a figure a bit bigger? The images are too small. Also please make the font size a bit bigger as well.}
    }
    \label{fig:sketchmap}
\end{figure}

Structured light systems have gained significant traction in various applications as an effective solution for range sensing \cite{batlle1998recent, salvi2004pattern, salvi2010state, wu2022twofrequency}. A monocular structured light system typically comprises a camera and projector, each with calibrated intrinsic and extrinsic parameters. By projecting pre-designed patterns into the 3D space, depth information is extracted by analyzing captured image(s) for deformed pattern(s). In structured light systems, classical matching algorithms strive to establish robust and accurate correspondence across multiple projected patterns, as depicted in ~\cref{fig:sketchmap}. 

In structured light systems, the 3D scanning process involves inherent trade-offs between the precision of the scan and the number of captured images or projected patterns. These trade-offs arise due to uncertainty factors originating from environmental and device-related influences. Augmenting the number of images enhances correspondence determination and yields promising outcomes. However, a higher count of projected patterns leads to extended acquisition times, making the system more susceptible to undesired motion and usage constraints. To sustain accuracy with fewer projected patterns, researchers have concentrated on strategies that embed richer information within a constrained pattern set~\cite{young2007viewpoint, ishii2007high, mirdehghan2018optimal, lin2016single, gupta2018ageometric, hall2001stripe}. These approaches incorporate intricately designed patterns encoding temporal or spatial features to mitigate matching uncertainty. Decoding these features from captured images establishes correspondences for depth estimation~\cite{koninckx2006real, kawasaki2008dynamic, sagawa2009dense, sagawa2011dense}. Yet, designing features resilient against unexpected influences and yielding dense depth maps remains a formidable challenge. Recent advancements in deep learning have prompted researchers to leverage deep neural networks for addressing these uncertainties~\cite{riegler2019connecting, johari2021depthinspace, qiao2022tide}. While these methods enable dense depth map generation, they necessitate extensive training datasets that significantly impact network performance. Constructing such datasets for structured light systems is an arduous endeavor, especially when accommodating diverse devices and pattern combinations.

In our research, we take a different approach to address the depth reconstruction challenge. Rather than designing or learning robust features for matching, we propose a generative framework founded on a neural implicit function to tackle these concerns. Specifically, we employ a neural network to formulate the signed distance field (SDF) of the target 3D scene. Through a fully differentiable rendering process, we generate images from the camera viewpoint. Establishing an end-to-end training procedure involves a direct loss between captured images and the rendered counterparts, leveraging prior knowledge of projected patterns. Upon network convergence, the 3D space's SDF becomes accessible, facilitating the extraction of the object surface and depth map. This neural implicit field is extensively utilized in research akin to neural radiance fields (NeRF) for view synthesis problems~\cite{mildenhall2020nerf}. NeRF-based approaches exhibit compelling image synthesis from passive camera views; however, challenges emerge in geometry recovery~\cite{wang2021neus, yariv2021volume, yan2021continual, niemeyer2022regnerf}. These challenges stem from the need to jointly estimate radiance and geometry fields based on the captured images. In structured light systems, the radiance field is predefined by projected patterns. Thus, our focus is solely on the geometry field, enabling robust high-quality geometry estimation.

% Our proposed framework is inspired by the research that utilizes the neural radiance fields (NeRF)~\cite{mildenhall2020nerf}, which has opened up a new line of research in view synthesis and 3D reconstruction~\cite{wang2021neus, yariv2021volume, yan2021continual, niemeyer2022regnerf}. Although NeRF-based methods can provide very promising rendered images and recovered geometries for passive cameras, they face significant challenges when transferred to our structured light system. First, the geometry extracted from the network implicit fields is often blurry, leading to ambiguity in the final depth map. Second, NeRF-based methods typically assumes multiple images captured from different viewpoints to provide sufficient constraints for training, whereas our system have fixed viewpoint. The lack of images from multiple viewpoints can hinder network convergence, resulting in unstable reconstructions.

For training the neural signed distance field, we leverage the projected light field to impose supplementary constraints. To accommodate our single-view camera setup, we streamline the network function and introduce a rendering process that incorporates the constraints posed by projected patterns during training. Furthermore, we introduce a surface color loss to enhance accuracy. Experimental findings demonstrate that even with a limited set of binary patterns (6 patterns) projected, our method achieves smooth object surfaces while preserving clear object boundaries. Notably, as the number of projected patterns increases, the network progressively enhances accuracy over a few additional iterations, thus effectively reducing computational costs. To the best of our knowledge, this is the first work that applies the neural implicit functions for depth map reconstruction in the context of monocular camera structured light systems.

In summary, our work makes the following contributions:
\begin{itemize}
    \item \textbf{Learning-based depth reconstruction}: Instead of traditional matching techniques, we introduce a learning framework that obviates the need for extra training datasets for depth reconstruction in structured light systems. 
    \item \textbf{Neural SDF with enhanced geometry estimation}: We apply neural SDF in the context of structured light systems, making high-quality geometry estimation possible.
    \item \textbf{Effective training with projected pattern constraints}: Our approach leverage the constraints posed by projected patterns to train the neural SDF, enhancing network convergence and optimizing geometry estimation performance.
\end{itemize}

\section{Related Work}

\textbf{Temporal-encoding structured light systems.} In the context of structured light systems for static scene reconstruction, temporal-encoding patterns find widespread use. These patterns allow unique decoding for each camera pixel. Various temporal coding techniques have been proposed, encompassing grid patterns\cite{lei2013design}, binary strips~\cite{scharstein2003high}, gray codes~\cite{posdamer1982surface, sundar2022single, weinmann2011multi, aliaga2008photogeometric}, phase measurements~\cite{wang20133dabsolute, zuo2013highspeed}, and fringe patterns~\cite{koninckx2006real, kawasaki2008dynamic, sagawa2011dense, taguchi2012motion}. Contemporary structured light systems exhibit commendable accuracy with sufficient projected patterns and resilience against various sources of noise. However, accuracy markedly degrades in few-shot scenarios where only a limited number of patterns can be projected. Researchers have endeavored to ameliorate this issue by refining coding strategies using specialized devices~\cite{sundar2022single, weinmann2011multi, aliaga2008photogeometric}, complex pattern structures~\cite{scharstein2003high, lei2013design}, or post-processing global optimizations~\cite{koninckx2006real, kawasaki2008dynamic, sagawa2009dense, sagawa2011dense}. Nonetheless, manual modeling of uncertain factors introduces unstable outliers in the estimated depth maps.

\textbf{Matching with learning-based methods in structured light systems.} Several studies have leveraged learning-based techniques to tackle the matching problem in structured light systems, exemplified by works like~\cite{ryan2016hyperdepth, fanello2017ultrastereo}. Moreover, deep networks have been utilized to predict disparity through image-pattern pair(s) within these systems~\cite{zhang2018activestereonet,riegler2019connecting, johari2021depthinspace}. However, these learning-based frameworks necessitate substantial training datasets. Regrettably, due to the distinct attributes of projected patterns and parameter configurations, there exists a dearth of publicly available benchmarks catering to structured light systems. While a proposed approach in~\cite{qiao2022tide} can undergo pretraining on synthetic data, it remains susceptible to the challenge of domain shifting when transitioning to real-world scenarios.

\textbf{Neural representations of 3D geometry.} In recent times, neural representations employing positional encoding have garnered significant attention across diverse tasks such as novel-view synthesis, generative modeling, and 3D reconstruction~\cite{mildenhall2020nerf, yariv2021volume, wang2021neus, long2023neuraludf, niemeyer2022regnerf, attal2021torf}. Unlike conventional forms like point clouds, volumes, or meshes, the neural representations of the 3D space are considered promising for handling complex geometries~\cite{park2019deepsdf}. Although a differentiable renderer for mesh has been proposed, it is theoretically challenging to handle occlusion and intricate shapes~\cite{kato2018neural,liu2019softrasterizer}. In contrast, various studies have introduced differentiable rendering processes for training neural networks, circumventing such limitations. Among these, the Neural Radiance Fields (NeRF) \cite{mildenhall2020nerf} has emerged as a simple yet promising paradigm, spawning several subsequent works~\cite{attal2021torf, meuleman2022floatingfusion, deng2022depthsupervisednerf, xiangli2022bungeenerf, deng2022depth, xu2022sinnerf}. A multitude of efforts have been directed at enhancing geometric performance within the NeRF framework~\cite{darmon2022improving, wang2021neus, long2023neuraludf, fu2022geoneus, long2022sparseneus}. Notably, NeuS\cite{wang2021neus} replaces the density field with a Signed Distance Field (SDF) function and adapts the volumetric rendering process to bridge the gap between SDF value and rendering weights during training. In our work, we adopt the same neural network structure used in NeuS~\cite{wang2021neus} and design a training framework to for high-quality geometry reconstruction in structured light systems. Specifically, we leverage prior knowledge in the form of radiance fields and optimize geometry fields exclusively, capitalizing on the predefined nature of projected patterns in structured light systems.
% this approach as the backbone and modify the training process to suit our settings. % Specifically, our approach involves fixed camera and active projection of light from the projector. 

\begin{figure*}
   \centering
   \includegraphics[width=0.9\linewidth]{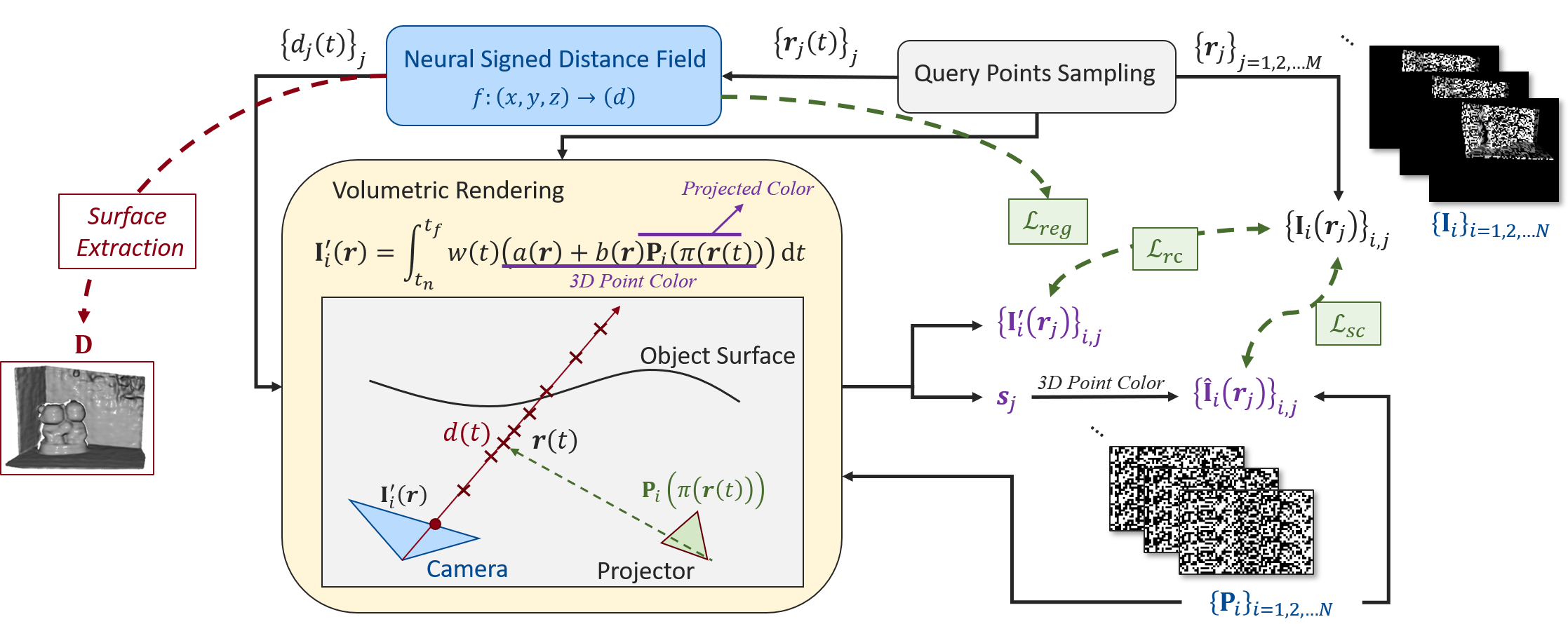}
   \caption{The general framework of our method for depth reconstruction using neural signed distance fields. Given a pixel from the camera space, the color/intensity is rendered using a combination of the neural signed distance field and the volumetric rendering process. The rendered color/intensity is then compared with the captured image to train the neural network.}
   \label{fig:framework}
\end{figure*}

\section{Methods}

In our approach, we employ a single camera and a projector to capture images for depth reconstruction. The projector projects multiple patterns, while the camera captures the images in turn. We represent the set of captured images as $\{\mathrm{\mathbf{I}}_i\}_{i=1,2,...N}$ and the set of projected patterns as $\{\mathrm{\mathbf{P}}_i\}_{i=1,2,...N}$, where $N$ represents the total number of projected patterns, as illustrated in ~\cref{fig:sketchmap}. The intrinsic and extrinsic parameters of both the camera and the projector are calibrated. Our objective is to reconstruct the depth map from the camera viewpoint, which we denote as $\mathrm{\mathbf{D}}$.

Traditional matching methods aim to discover a function to produce the desired depth map from the given input data. In contrast, we propose a generative framework in our paper, as depicted in ~\cref{fig:framework}. Firstly, we employ a neural network to represent the signed distance field, as illustrated in \cref{sec:NeuSDF}. Subsequently, a differentiable process is employed for volumetric rendering of the camera-observed image, as presented in \cref{sec:volumericrendering}. Finally, the network is trained using a loss function, which involves a direct comparison of the rendered images with the captured ones, as outlined in \cref{sec:lossfunction}. 

\subsection{Neural Signed Distance Field}\label{sec:NeuSDF}

We represent the 3D scene as an implicit function whose input is a 3D location $\mathbf{x}=\left(x,y,z\right)$ and whose output is the signed distance value $d$. This implicit function can be defined as $f$, where

\begin{equation}
    f: \left(x,y,z\right) \rightarrow d .
\end{equation}

The surface of the object is represented by the zero-level set of the signed distance function $f$ above. This representation of the 3D scene is akin to the one presented in NeuS~\cite{wang2021neus}, but more straightforward and uncomplicated. We leverage the same positional encoding embedders and Multilayer Perceptron (MLP) network as proposed in NeuS, while excluding the viewing direction input as our viewpoint remains fixed. Additionally, we do not estimate the emitted color since we observe that the distance value itself is enough for image rendering in structured light systems. 

\subsection{Volumetric Rendering with Projected Patterns}\label{sec:volumericrendering}

We denote one of the rendered images as $\mathrm{\mathbf{I}}'_i$. For each pixel $\mathbf{r}$ in the rendered image, we compute the expected color $\mathrm{\mathbf{I}}'_i(\mathbf{r})$ using classic volume rendering~\cite{kajiya1984ray} with the near and far bounds $t_n, t_f$ along the light ray:

\begin{equation}\label{eq:rendering}
    \mathrm{\mathbf{I}}'_i(\mathbf{r}) = \int_{t_n}^{t_f}
    w(t) \mathbf{c}(\mathbf{r}(t)) \mathrm{d}t ,
\end{equation}
where $\mathbf{r}(t) = t\mathbf{v}$ is the projected light ray with the direction vector $\mathbf{v}$. The weight $w(t)$ for integration represents the likelihood that point $\mathbf{r}(t)$ lies on the surface. In NeuS~\cite{wang2021neus}, an occlusion-aware weight function is constructed using the estimated SDF value:

\begin{equation}
    w(t) = \frac{
        \phi_S(f(\mathbf{r}(t)))
    }
    {
        \int_{0}^{+\infty}\phi_S(f(\mathbf{r}(u))) \mathrm{d}u
    }.
\end{equation}

$\phi_S$ is the logistic density distribution:

\begin{equation}
    \phi_S(x) = \frac{se^{-sx}}{(1 + e^{-sx})^2},
\end{equation}
where $s$ is a trainable parameter and  $1/s$ approaches to zero as the network training converges. Further details of the weight function $w(t)$ can be found in~\cite{wang2021neus}.

In our approach, the color of the 3D point $\mathbf{r}(t)$ is not explicitly estimated from the network. Instead, we leverage the prior knowledge of an active light field to compute the color projected onto the 3D point via a reprojection function $\pi$. Specifically, the color $\mathbf{c}\left(\mathbf{r}(t)\right)$ of the 3D point can be obtained from the projected pattern $\mathrm{\mathbf{P}}_i$ in our case, through:

\begin{equation}
    \mathbf{c}\left(\mathbf{r}(t)\right) =
    a(\mathbf{r}) + b(\mathbf{r}) \mathrm{\mathbf{P}}_i\left(
        \pi(\mathbf{r}(t))
    \right) ,
\end{equation}
where $\mathrm{\mathbf{P}}_i(\pi(\mathbf{r}(t)))$ symbolized the projected color calculated through the reprojection operation based on the intersection between projected light and volume in space. Subsequently, this projected color contributes to the estimation of the 3D point's color employing a linear model. In the 2D camera space, $a(\mathbf{r})$ and $b(\mathbf{r})$ denote the background light level and fringe contrast, respectively. These two parameters are derived from the captured images:

\begin{equation}
    \left\{
        \begin{array}{ll}
             a(\mathbf{r}) = \mathrm{min}(\{\mathrm{\mathbf{I}}_i(\mathbf{r})\}_{i=1,2,...N}) & \\
             b(\mathbf{r}) = \mathrm{max}(\{\mathrm{\mathbf{I}}_i(\mathbf{r})\}_{i=1,2,...N}) - a(\mathbf{r}) &
        \end{array}
    \right. .
\end{equation}

It is noteworthy that these two parameters inherently facilitate the masking of occluded regions, where obstructed projected light rays encounter objects. This is achieved as the fringe contrast and background light level tend to approach zero within those areas.

\begin{figure}
    \centering
    \includegraphics[width=0.9\linewidth]{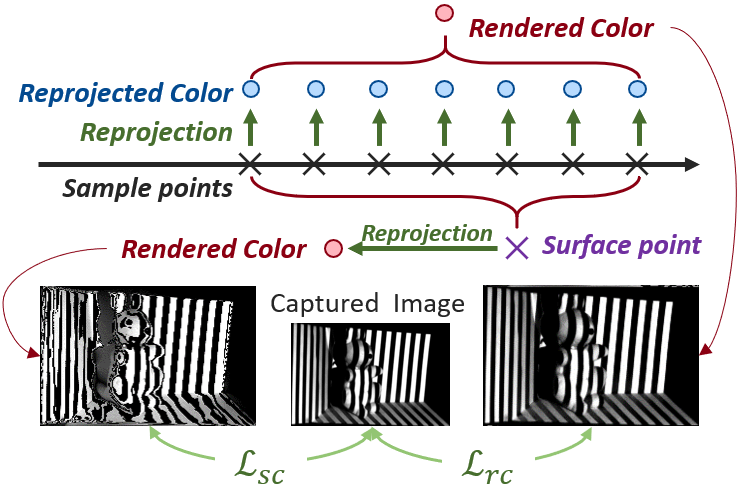}
    \caption{Distinguishing between Rendered Color Loss $\mathcal{L}_{rc}$ and Surface Color Loss $\mathcal{L}_{sc}$.  $\mathcal{L}_{sc}$ enforces a singular peak along the projection ray for geometry consistency, while $\mathcal{L}_{rc}$ emphasizes accurate color in rendered images. In an example, correct rendered color (upper) contrasts with flawed surface color (lower), highlighting geometry inaccuracies in wave-like regions.
    % \knote{Fonts are a bit too small, too.}
    }
    \label{fig:colorloss}
\end{figure}

\subsection{Loss Functions}\label{sec:lossfunction}

During the training process, a batch of $M$ pixels, represented as $\{\mathbf{r}_j\}_{j=1,2,...M}$, is randomly selected from the captured image. The rendered colors of these selected pixels are then estimated, and the network is trained by minimizing a per-pixel loss function that quantifies the difference between the rendered colors and the captured colors. In our work, the loss function is defined as

\begin{equation}
    \mathcal{L} = \mathcal{L}_{rc} 
    + \lambda_{sc} \mathcal{L}_{sc} 
    + \lambda_{reg} \mathcal{L}_{reg} ,
\end{equation}
where $\lambda_{*}$ is the leverage parameters for additional terms. The rendered color loss $\mathcal{L}_{rc}$ directly constraints the rendered images:

\begin{equation}
    \mathcal{L}_{rc} = \frac{1}{MN}\sum_{j=1}^{M} \sum_{i=1}^{N} \left|
        \mathrm{\mathbf{I}}_i(\mathbf{r}_j)
        - \mathrm{\mathbf{I}}'_i(\mathbf{r}_j)
    \right|_{1} .
\end{equation}

The $\mathcal{L}_{reg}$ is a regular terms conducted in NeuS~\cite{wang2021neus}. The Eikonal term $\mathcal{L}_{reg}$ constraints the smooth distribution of the neural field:

\begin{equation}
    \mathcal{L}_{reg} = \frac{1}{M(t_f - t_n)} \sum_{j=1}^M \int_{t_n}^{t_f} \left(
        \left| \nabla f(\mathbf{r}_j(t)) \right|_2 - 1
    \right)^2 \mathrm{d}t .
\end{equation}

In practice, it has been observed that when computing the weight $w(t)$ from the SDF, there may be multiple peaks along the projecting ray to produce a precise color, resulting in lots of floaters in the reconstructed geometry shape. This effect can be particularly pronounced in our single-view experiment settings. To address this issue, we additionally apply another color constraint to the final extracted surface. We first compute the expected surface point~\cite{niemeyer2022regnerf} $\mathbf{s}_j$ for the light ray $\mathbf{r}_j(t)$ as:

\begin{equation}\label{eq:surface}
    \mathbf{s}_j = \int_{t_n}^{t_f}
        w(t)\mathbf{r}_j(t)
    dt .
\end{equation}

Then the pixel color of the expected surface point can be computed from the warping function $\pi$:

\begin{equation}
    \mathrm{\mathbf{\hat{I}}}_i(\mathbf{r}_j) =
    a(\mathbf{r}_j) + b(\mathbf{r}_j)
    \mathrm{\mathbf{P}}_i(\pi(\mathbf{s}_j)) .
\end{equation}

$\mathrm{\mathbf{\hat{I}}}_i$ is the rendered color from the expected surface point. We formulate the surface color loss $\mathcal{L}_{sc}$ as

\begin{equation}
    \mathcal{L}_{sc} = \frac{1}{MN}\sum_{j=1}^{M} \sum_{i=1}^{N} \left|
        \mathrm{\mathbf{I}}_i(\mathbf{r}_j)
        - \mathrm{\mathbf{\hat{I}}}_i(\mathbf{r}_j)
    \right|_{1} .
\end{equation}

The surface color loss used in our approach shares similarities with photometric constraints commonly utilized in matching-based techniques~\cite{garg2016unsupervised, zhang2018activestereonet}. These constraints involve warping color from one image into another based on pixel depth information. While maintaining photometric consistency in novel view synthesis problems poses challenges due to view-dependent color variations, it can enhance geometry performance and improve final outcomes, especially in few-shot scenarios~\cite{niemeyer2022regnerf, xiangli2022bungeenerf, darmon2022improving}. Our focus prioritizes geometry performance over photorealism. Thus, enforcing photometric consistency aids the network in generating superior 3D shapes without introducing rendering ambiguities. Additionally, we provide a sketch map to illustrate the contrast between rendered color loss and surface color loss in \cref{fig:colorloss}.

Upon network convergence, the depth map is derived by computing the anticipated surface through the implicit signed distance function using the marching cube algorithm. Concretely, the surface is reprojected into camera space to compute the depth map $\mathrm{\mathbf{D}}$.

\textbf{Incremental optimization.} Our framework seamlessly accommodates additional input patterns during optimization. In an online scenario, after optimizing the network for several iterations, introducing a new pattern-image pair does not necessitate starting the optimization process from scratch. Instead, we increment the total frame count from $N$ to $N+1$ and seamlessly resume the optimization. The network leverages the new constraints from the updated pattern sets. This approach eliminates redundant computations and enhances computing efficiency in incremental cases.
% Our framework is friendly to the additional input patterns during optimization. Considering an online case that the network has been optimized for several iterations and we have a new pattern-image pair as additional input to this framework. We do not need to optimize the network from the beginning. Just increase the total frame number $N$ to $N+1$ and continue the optimizing process and the network can utilize the new constraints from the pattern sets. Therefore we can avoid the repeating computation and improve the computing time performance for incremental case.

\begin{table*}
    \begin{center}
        \begin{tabular}{@{}ccccccc@{}}
            \hline
            Methods & Scene 1 & Scene 2 & Scene 3 & Scene 4 & Scene 5 & Scene 6 \\
            \hline
            N-PMP (6)~\cite{wang20133dabsolute} & 529.966 & 533.876 & 599.681 & 516.122 & 537.898 & 527.607 \\
            H-PMP (6)~\cite{zuo2013highspeed} & 14.973 & 19.082 & 48.467 & 17.272 & 41.117 & 40.795 \\
            CGC (7)~\cite{wu2020highspeed} & 34.135 & 27.120 & 10.660 & 30.643 & 40.337 & 37.196 \\
            CGC (6)~\cite{wu2020highspeed} & 45.699 & 44.782 & 91.433 & 39.009 & 71.247 & 52.945 \\
            GC (7)~\cite{posdamer1982surface, aliaga2008photogeometric, weinmann2011multi} & 8.277 & 7.689 & 6.584 & 5.391 & 10.961 & 7.225 \\
            GC (6)~\cite{posdamer1982surface, aliaga2008photogeometric, weinmann2011multi} & 15.766 & 10.498 & 6.672 & 9.626 & 16.333 & 10.177 \\
            Ours (6) & \textbf{4.279} & \textbf{3.177} & \textbf{3.103} & \textbf{2.653} & \textbf{2.985} & \textbf{3.204} \\
            \hline
        \end{tabular}
    \end{center}
    \caption{The experiment results compared with the classic methods. The number of patterns used for each method is marked in parentheses. Note that the lower the number, the better the performance.}
    \label{tab:mainexp}
\end{table*}

\begin{figure}
   \centering
   \includegraphics[width=\linewidth]{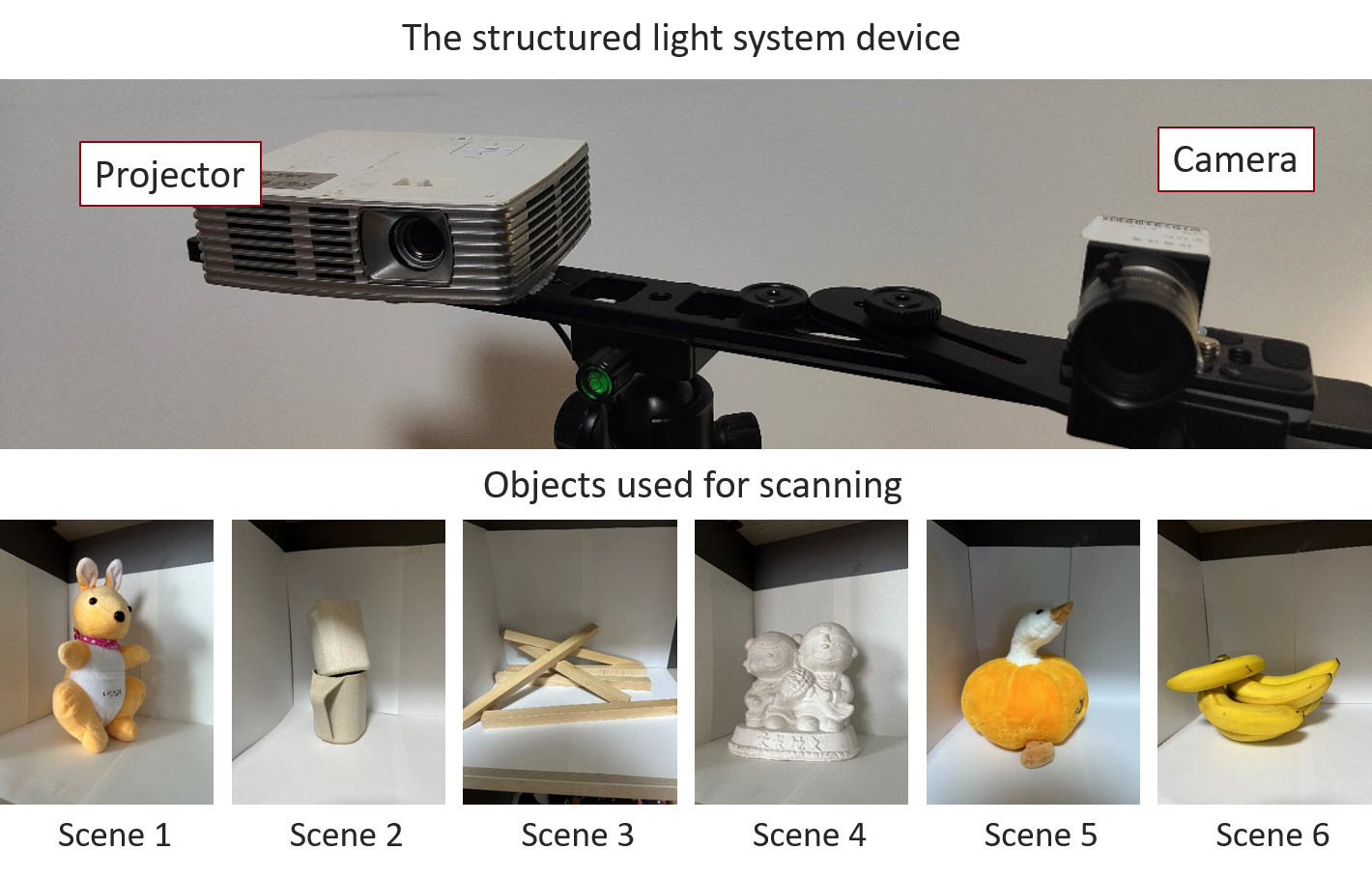}
   \caption{The experiments setting for our data collection. Six scenarios with different objects are used for our data collection.}
   \label{fig:device}
\end{figure}

% \begin{figure}
%    \centering
%    \includegraphics[width=0.9\linewidth]{figures/patterns.png}
%    \caption{Patterns utilized by comparison results.}
%    \label{fig:patterns}
% \end{figure}

\begin{table*}
  \centering
  \begin{tabular}{@{}ccccccc@{}}
    \hline
    Losses & Scene 1 & Scene 2 & Scene 3 & Scene 4 & Scene 5 & Scene 6 \\
    \hline % 3 pattern
    $\mathcal{L}_{rc}$ & 19.739 & 19.227 & 12.267 & 11.417 & 6.242 & 13.033 \\
    $\mathcal{L}_{rc}, \mathcal{L}_{reg}$ & 14.973 & 8.825 & 7.040 & 4.868 & 5.281 & 5.864 \\
    $\mathcal{L}_{sc}$ & 74.74 & 754.943 & 737.504 & 579.428 & 764.962 & 745.885 \\
    $\mathcal{L}_{sc}, \mathcal{L}_{reg}$ & 782.529 & 754.943 & 109.536 & 741.676 & 764.962 & 745.885 \\
    % $\mathcal{L}_{rc}, \mathcal{L}_{sc}$ & xx & xx & xx & xx & xx & xx \\
    $\mathcal{L}_{rc}, \mathcal{L}_{sc}, \mathcal{L}_{reg}$~(Ours) & \textbf{4.279} & \textbf{3.177} & \textbf{3.103} & \textbf{2.653} & \textbf{2.985} & \textbf{3.204} \\
    \hline
  \end{tabular}
  \caption{Ablation studies conducted to assess the impact of the rendered color loss and the surface color loss. While the surface loss $\mathcal{L}_{sc}$ ensures a more accurate geometry shape, the one-point constraint could potentially lead to divergence in the optimization process. As a remedy, we amalgamate these two losses to ensure convergence.}
  \label{tab:ablation}
\end{table*}

\section{Experiments}

\subsection{Hardware Platform}
Our structured light system comprises a single grayscale camera and a projector, depicted in~\cref{fig:device}. All intrinsic and extrinsic parameters are pre-calibrated. We collect data from six different scenes with varying objects for comparison using our platform, as depicted in the same figure. To establish the ground truth data for comparison, we project an additional 7 gray code patterns with inverse color for decoding, along with four phase-shifting patterns~\cite{kajiya1984ray} to generate sub-pixel level depth maps. These patterns serve exclusively for generating accurate reference data~\cite{scharstein2003high}. Post depth map acquisition, we manually eliminate outlier noise and generate a boundary mask for visualization purposes.

\begin{figure*}
   \centering
   \includegraphics[width=0.9\linewidth]{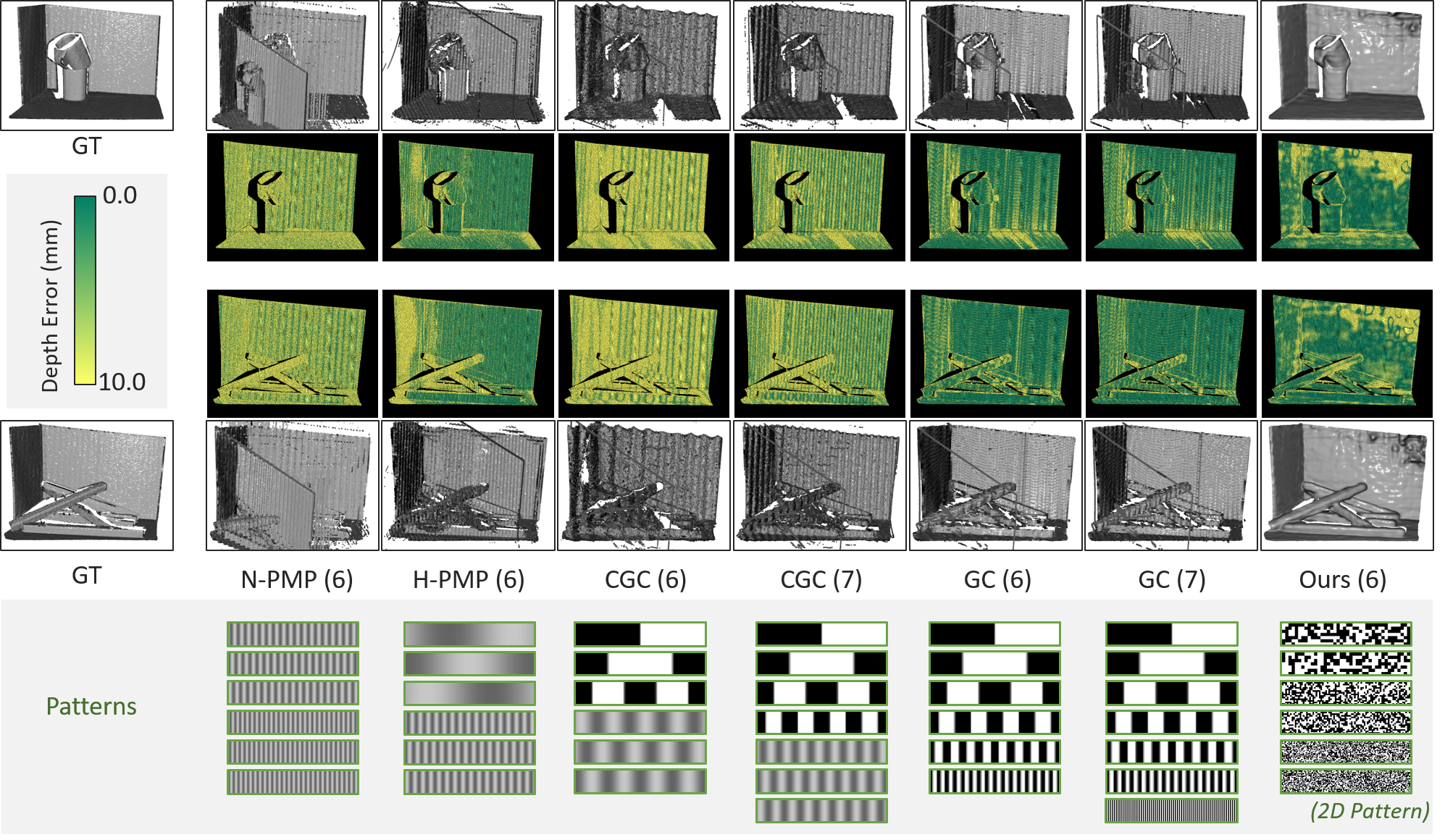}
   \caption{Visualization of error maps and point clouds in two scenes. We also append the patterns used by each method at the bottom of each method.}
   \label{fig:mainexp}
\end{figure*}

\begin{figure*}
   \centering
   \includegraphics[width=\linewidth]{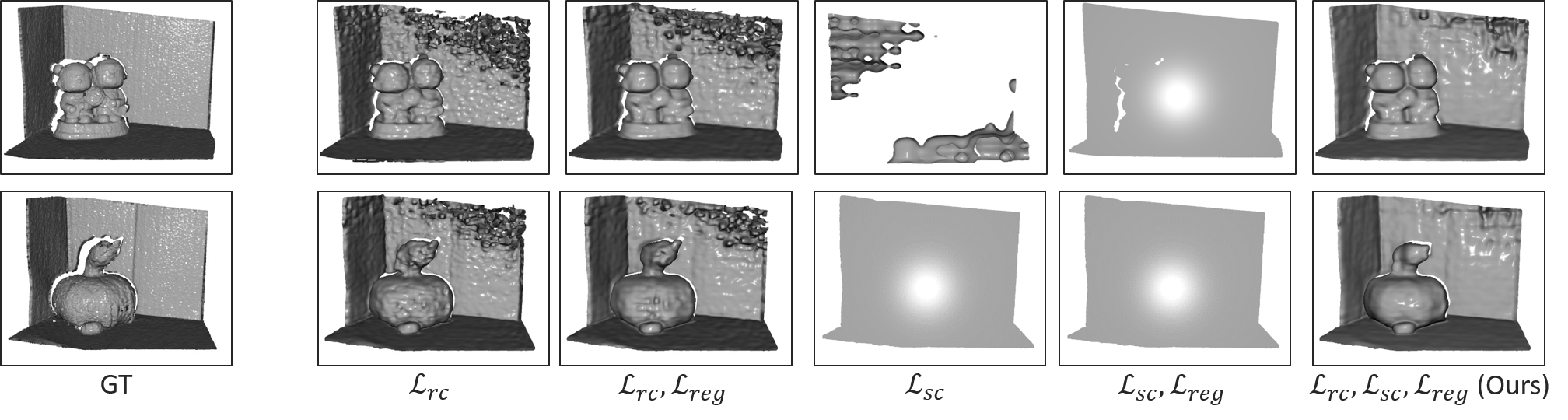}
   \caption{Ablation studies of the loss functions.}
   \label{fig:ablation}
\end{figure*}

\subsection{Pattern Sets}\label{sec:patterndesign}
Our framework eliminates the need for explicit encoding-decoding techniques embedded with the pattern sets. Instead, patterns serve as color constraints for training the neural implicit field in the 3D space. As such, we furnish our framework with a collection of randomly generated array-based patterns. The projector space is subdivided into uniform small squares, each randomly assigned black or white colors. Employing a multi-scale configuration, we employ unit squares of length 20,10, and 5 pixels for each scale. In the final step, a total of 6 patterns are utilized (two patterns per scale), forming the basis for a series of comparative experiments. % The supplementary materials include generated random patterns.  % TODO: Patterns

\subsection{Comparisons}
To train the network, we initiate 1,000 iterations with $\lambda_{sc}=0$ and $\lambda_{reg}=0.1$. The omission of surface color loss in the initial training phase is intended to circumvent potential issues related to local minima. Subsequently, we modify the parameters to $\lambda_{sc}=1.0$ and continue training for an extra 3,000 iterations. The complete training process consumes approximately 14 minutes using a single RTX2080Ti GPU card. % Further implementation specifics can be found in the supplementary materials. % TODO: Details

We compare our method to four classic decoding methods because decoding-based structured light methods have not changed essentially these 10 years,
which require similar pattern numbers for decoding: The hierarchical phase measurement profilometry (H-PMP)~\cite{wang20133dabsolute}, the numerical phase measurement profilometry (N-PMP)~\cite{zuo2013highspeed}, the complementary gray code (CGC)~\cite{wang2011optics, wu2020highspeed}, and the binary gray code (GC)~\cite{posdamer1982surface, aliaga2008photogeometric, weinmann2011multi} with interpolation between fringes. It is worth noting that NeRF-based approaches~\cite{wang2021neus, niemeyer2022regnerf, xiangli2022bungeenerf, long2023neuraludf, xu2022sinnerf} could not make any meaningful results because they could not handle active lights like structured light, so they are excluded from the experiments. H-PMP and N-PMP require six patterns, while the CGC and GC require six or seven patterns. The patterns used by each method are displayed at the bottom, shown in ~\cref{fig:mainexp}. We mark the pattern numbers used by each method in parentheses. Our method only requires six binary code patterns for training. By applying the marching cube algorithm, we derive the mesh, and then reproject it into the camera space to generate the depth map. Quantitative assessment of the estimated depth maps is conducted via the average L1 loss. Experiment results are presented in ~\cref{tab:mainexp}. For better visualization, error maps and point clouds, computed from the estimated depth maps, are included in ~\cref{fig:mainexp}.

The H-PMP, N-PMP, and CGC use phase-shifting encoding, which encodes the coordinates into a set of sinusoidal patterns. The periodic natures of the sinusoidal patterns cause phase ambiguity during decoding, which can be resolved by projecting additional patterns, as done in the ground truth acquisition process. When the wavelength is within a short range (around 40 pixels), phase shifting can achieve high precision. However, when the pattern number is limited, resolving this problem becomes challenging, as only three or four additional patterns can be utilized for decoding. H-PMP uses another set of phase patterns with a wavelength equal to the image width, but this method is susceptible to shading factors. N-PMP uses two sets of phase patterns with a relatively smaller wavelength, but the method is sensitive to the error of the decoded phase value. CGC uses binary gray-code patterns for decoding, but up to four patterns require a minimum wavelength of 160 pixels, leading to a loss of accuracy. GC relies solely on binary gray codes for decoding, potentially leading to accuracy loss between projected binary fringes. In contrast, our method employs six binary patterns without continuous-encoded patterns, yet the network achieves a smooth surface. 

% In addition, we also examine two possible implementations based on the neural implicit functions. While vanilla NeRF~\cite{mildenhall2020nerf, niemeyer2022regnerf} uses density fields for 3D representation, many papers focusing on geometric reconstruction prefer signed distance fields~\cite{wang2021neus, yariv2021volume}. To explore this, we implement a version based on the density fields with our rendering processing and loss function training, which we refer to as 'Density-based' in ~\cref{tab:ablation} and ~\cref{fig:ablation}. Besides, we also conduct the OnSurfacePrior, who learns a neural SDF from the point cloud. We feed the point cloud estimated from "GC (7)" into this method and evaluate the results. Our results clearly show that our representation achieves a better shape in our structured light systems.

\subsection{Analysis}
To assess the impact of different combinations of loss functions in our proposed framework, we conduct ablation studies (\cref{tab:ablation}, \cref{fig:ablation}). Results reveal that while the surface color loss enhances overall shape and geometry preservation, relying solely on it can lead to suboptimal local minima and inadequate reconstruction. Thus, the rendered color loss remains crucial in our framework. Additionally, the Eikonal loss as a regularization term contributes to smoother object surfaces and improved final performance.

\begin{figure*}
   \centering
   \includegraphics[width=0.95\linewidth]{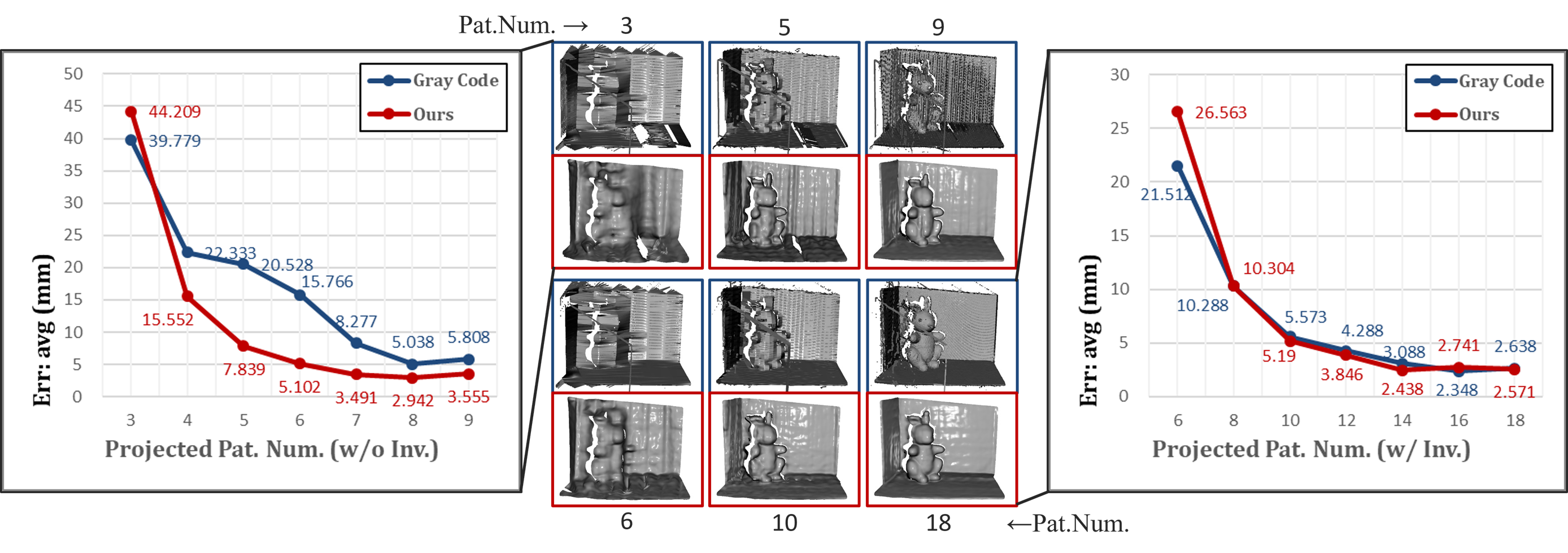}
   \caption{Comparison with gray code using the same pattern sets. We conduct a performance evaluation using the full Gray code pattern set, employing up to 9 patterns, for both classical methods and our learning-based framework. The chart illustrates network performance across varying pattern numbers. Due to the limitation of space, we only present the result based on one scene. }  % Please refer to the supplementary materials for more experimental results.
   \label{fig:graycode}
\end{figure*}

In order to effectively demonstrate the prowess of our learning-based framework in comparison to classical matching algorithms, we carry out further experiments using Gray code methods with the same patterns for both. We implement both fixed threshold (w/o Inv.) and inverse projection (w/ Inv.) algorithms for pixel recognition, following Dong Lanman's tutorial~\cite{lanman2009build}. Employing up to 9 projected patterns with an interval of 2.5 pixels, we achieve superior accuracy compared to Gray code methods with the same patterns (\cref{fig:graycode}). Notably, in scenarios where only 3 patterns are available, the conventional method exhibits better performance due to its ability to fit the background wall using linear interpolation. 

\begin{figure*}
   \centering
   \includegraphics[width=0.9\linewidth]{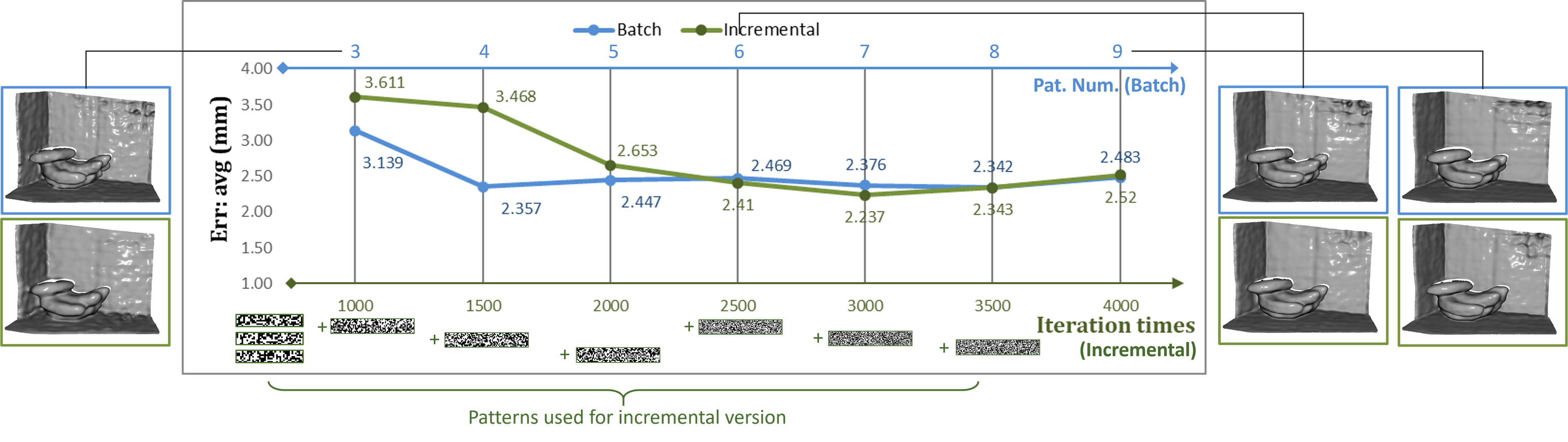}
   \caption{Batch vs. Incremental training comparison. We employ distinct colors to depict the loss and mesh for visualization. The pattern sets utilized for the incremental version are showcased at the bottom of the chart.}
   \label{fig:online}
\end{figure*}

Furthermore, we evaluate accuracy and convergence speed with varying pattern numbers within our framework. Employing the pattern set introduced in \cref{sec:patterndesign}, we increase the pattern count to 9, with 3 patterns per scale. In the batch computing version, the network is trained for 4,000 iterations with varying input pattern numbers, and performance is evaluated at the final output. For the incremental version, simulating the online application, we initiate the first 1,000 iterations with 3 input patterns, progressively adding one pattern per 500 iterations until all patterns were utilized. Accuracy is assessed at the end of each 500-iteration interval, representing the output during online usage. The results, shown in \cref{fig:online}, demonstrate increased accuracy with more projected patterns for both 'Batch' and 'Incremental' versions. While the incremental approach initially lagged behind the final batch output, it provides a discernible general shape. With more iterations and patterns input, the incremental framework exhibit performance comparable to the batch version within a few hundred iterations. This nature of our framework significantly reduces computational costs by obviating optimization from the initial state during online applications.

\section{Conclusion}

In this paper, we introduce an innovative framework for depth reconstruction within structured light systems, centered around neural signed distance fields. Our framework revolves around training a neural implicit function to represent the scene's signed distance field. Employing a fully differentiable rendering process, the network is trained using the constraints from projected patterns. Following convergence, the depth map is retrieved by reprojecting the surface derived from the neural SDF. Experimental results reveal our method's effectiveness with minimal projected patterns. Furthermore, as pattern numbers increase, our method outperforms classical techniques and remains amenable to the incremental optimization process. To the best of our knowledge, this is the pioneer exploration of neural implicit functions within structured light systems. Our code will be accessible on our project website shortly.

The primary objective of this paper is to explore the potential of neural implicit representations within structured light systems. Although the obtained results are promising, several limitations persist, necessitating further research. Notably, the significant computational resources and minutes of training times required by our approach may limit its applicability in certain scenarios when compared to traditional decoding methods. Challenges such as wave-like effects on the object surfaces and unstable scene boundaries persist, particularly in few-shot cases. Furthermore, our framework's absence of explicit encoding-decoding features embedded in projected patterns suggests that an interesting future research topic would be to analyze optimal patterns for the neural implicit fields in structured light systems. We plan to address these limitations and explore this topic further in our future work. 

\textbf{Acknowledgement}. This work was supported by the National Natural Science Foundation of China (62176010), the Joint Funds of the National Natural Science Foundation of China (U22A2061) and JSPS KAKENHI Grant Number 
JP20H00611, JP21H01457, JP23H03439 in Japan.
{
    \small
    \bibliographystyle{ieeenat_fullname}
    \bibliography{main}
}

\end{document}